\documentclass[conference]{IEEEtran}

\IEEEoverridecommandlockouts  
\usepackage{cite}
\usepackage{amsmath,amssymb,amsfonts}
\usepackage{algorithmic}
\usepackage{graphicx}
\usepackage{textcomp}
\usepackage{xcolor}
\usepackage{colortbl}
\usepackage{array}
\usepackage{hyperref}
\usepackage{enumitem}
\usepackage{multirow}
\usepackage{tabulary}
\usepackage[section]{placeins}

\usepackage{float}
\usepackage[utf8]{inputenc}
\DeclareUnicodeCharacter{2002}{\ }
\usepackage{etoolbox}

\usepackage{geometry}
\usepackage{fancyhdr}
\renewcommand{\vec}[1]{\mathbf{#1}}

\title{\LARGE \bf
	LoopDB: A Loop Closure Dataset for Large Scale Simultaneous Localization and Mapping
}

\author{Mohammad-Maher Nakshbandi, Ziad Sharawy, Dorian Cojocaru, Sorin Grigorescu
	\thanks{Mohammad-Maher Nakshbandi, Ziad Sharawy and Sorin Grigorescu are with the Robotics, Vision and Control Laboratory (RovisLab, \url{https://www.rovislab.com}), Transilvania University of Brasov, Romania. Dorian Cojocaru is with the Department of Mechatronics and Robotics, University of Craiova. {\tt\small mohammad.nakshbandi@unitbv.ro}}%
}

\begin{document}
	\maketitle
	
	\begin{abstract}
		In this study, we introduce LoopDB, which is a challenging loop closure dataset comprising of over $1000$ images captured across diverse environments, including parks, indoor scenes, parking spaces, as well as centered around individual objects. Each scene is represented by a sequence of five consecutive images. The dataset was collected using a high resolution camera, providing suitable imagery for benchmarking the accuracy of loop closure algorithms, typically used in simultaneous localization and mapping. As ground truth information, we provide computed rotations and translations between each consecutive images. Additional to its benchmarking goal, the dataset can be used to train and fine-tune loop closure methods based on deep neural networks. LoopDB is publicly available at \url{https://github.com/RovisLab/LoopDB}.
	\end{abstract}
	
	
	\section{Introduction}
	
	Automous robots rely on simultaneous localization and mapping (SLAM) to build an environment map and pinpoint its position on it ~\cite{Mur-Artal2015ORBSLAM}. In particular, loop closure is a crucial component of SLAM that helps correct drift and improve the accuracy of the generated map. As a robot or autonomous vehicle explores an environment, small errors in odometry and sensor measurements accumulate over time, leading to inconsistencies in the estimated trajectories. Loop closure \cite{Tsintotas_2022}, \cite{cummins2008fabmap} detects when the system revisits a previously mapped location, allowing for the correction of past pose estimates through optimization techniques, such as pose graph optimization or bundle adjustment.
	
	Despite the critical importance of loop closure for accurate SLAM,  the existing datasets have significant limitations in the development of robust loop-closure algorithms. Existing datasets suffer from either restricted viewpoint diversity, limited scene diversity, or vague transformation metadata between loop-closure instances. These limitations prevent the development of viewpoint-invariant loop closure detection approaches for seamless operation in various real-world scenarios. LoopDB specifically addresses these challenges using  a set of tightly controlled multi-angle captures of the same scene, accurate transformation metadata, and high scene diversity across both indoor and outdoor environments.
	
	Autonomous robot development relies heavily on a large amount of real-world data for experimenting with, developing, and testing algorithms before deployment in real-life scenarios. The computer vision community has long favored a benchmark-driven approach \cite{nister2006visual}, and more recently, various vision-based autonomous robotics datasets have been made publicly available \cite{dollar2009pedestrian} \cite{brostow2009semantic} \cite{pandey2011ford} \cite{pfeiffer2013exploiting} \cite{blanco2014malaga} such as the KITTI dataset \cite{geiger2012kitti} and the Euroc dataset in \cite{burri2016euroc}.
	
	\begin{figure}
		\centering
		\includegraphics[width=0.9\columnwidth]{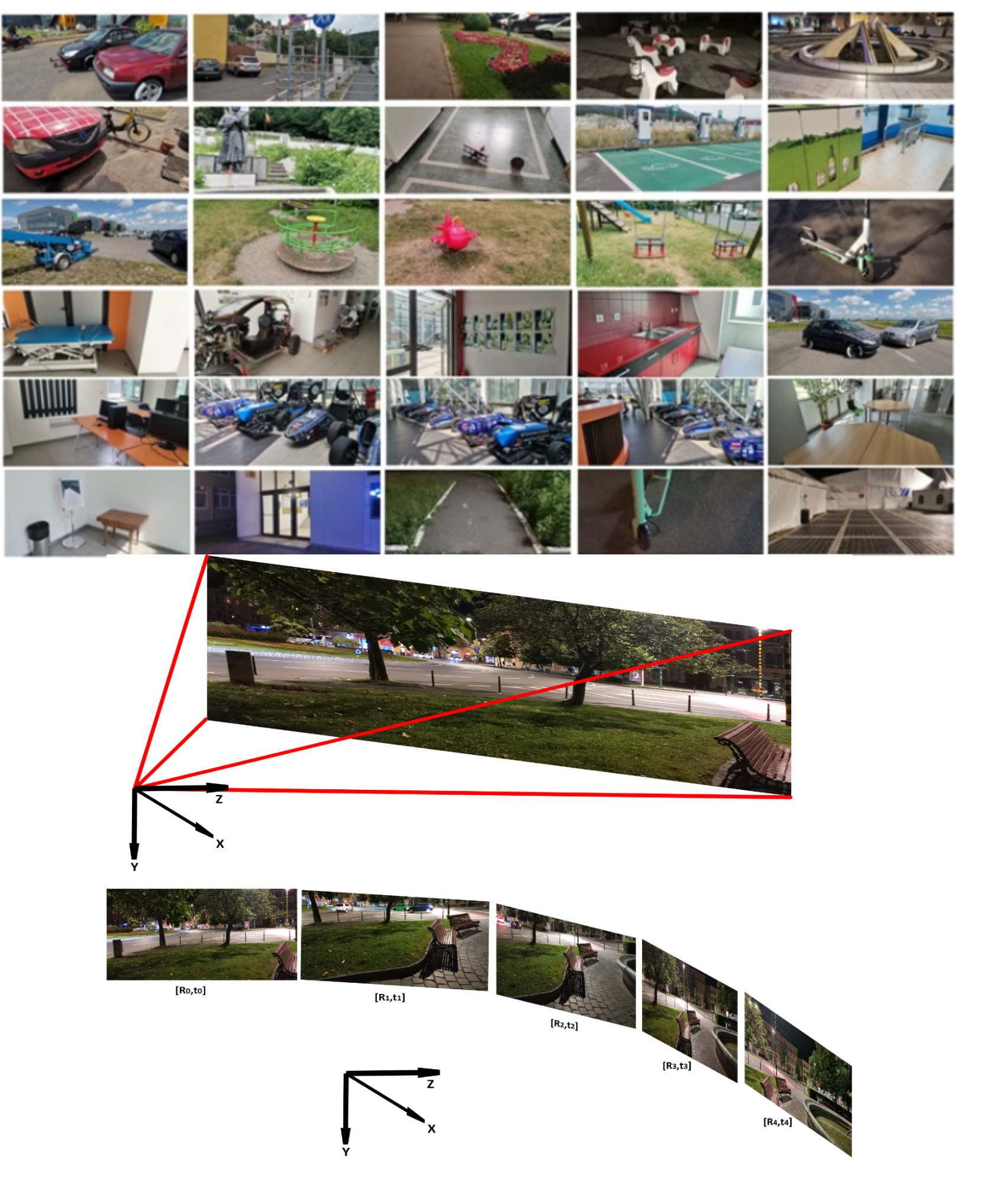}
		\caption{\textbf{LoopDB samples (top) and data acquisition procedure (bottom).} Each LoopDB sequence consists of five consecutive images, having their rotations and translations given as ground truth.}
		\label{fig:intro}
	\end{figure}
	
	LoopDB takes a different approach from existing datasets by focusing on the loop closure problem rather than on general SLAM or navigation. Our methodology involves  capturing multiple views (five per scene) per scene at over 200 different locations with precise camera pose information, enabling researchers to test and evaluate loop closure algorithms under controlled but challenging viewpoint variations. Unlike trajectory-based datasets, where loop closures occur randomly, LoopDB provides guaranteed loop closures with known transformations, allowing for quantitative evaluation of detection accuracy and geometric consistency. This allows the development of benchmark loop closure algorithms that can handle significant viewpoint changes, which is a common problem in real-world robotic applications that are not well addressed by existing datasets.

	\begin{table*}[!t]
		\renewcommand{\arraystretch}{1.2}
		\caption{Comparative Analysis of Loop Closure Detection Datasets}
		\label{table:comparison}
		\centering
		\resizebox{\textwidth}{!}{%
			\begin{tabular}{|p{1.8cm}|p{2.5cm}|p{2.5cm}|p{2.5cm}|p{2.5cm}|p{2.5cm}|p{2.5cm}|}
				\hline
				\textbf{Characteristic} & \textbf{LoopDB (Ours)} & \textbf{KITTI \cite{geiger2012kitti}} & \textbf{Oxford RobotCar \cite{maddern2017robotcar}} & \textbf{EuRoC MAV \cite{burri2016euroc}} & \textbf{4Seasons \cite{wenzel20204seasons}} & \textbf{KITTI-360 \cite{liao2022kitti}} \\
				\hline
				\textbf{Environment Types} & 
				Indoor: Gyms, Dorms, Hotels. Outdoor: Parks, Streets, Bus stations. & 
				Urban and rural roads in Karlsruhe & 
				Oxford city streets with varying conditions &
				Indoor industrial, Machine hall, Vicon room &
				Urban roads across all four seasons &
				360° view of urban and suburban environments \\
				\hline
				\textbf{Image Resolution} & 48MP (Huawei Nova 7i) & 1.4MP stereo cameras & High-res stereo cameras & 752×480 stereo cameras & 1920×1200 stereo & 1408×1408 panoramic + stereo \\
				\hline
				\textbf{Ground Truth} & Visual alignment with transformation metadata & GPS/IMU trajectory data & GPS/INS with weather data & 
				Vicon motion capture, Leica MS50 position & RTK-GPS, LiDAR maps & High-precision GPS/IMU, 3D semantic labels \\
				\hline
				\textbf{Primary Use Case} & Loop closure detection, Visual SLAM & Visual odometry, SLAM & Long-term autonomy, Place recognition & Visual-inertial odometry, MAV navigation & Long-term localization, Seasonal robustness & 360° perception, Semantic mapping \\
				\hline
				\textbf{Dataset Size} & ~1000 images (5 per scene) & 39,000+ images & 1M+ frames spanning 1 year & 230,000+ stereo frames & 318,000+ frames & 320,000+ images with 100km trajectory \\
				\hline
				\textbf{Sensor Suite} & 
				Single camera & 
				Stereo cameras, GPS, LiDAR & 
				Stereo cameras, LiDAR, GPS/INS &
				Stereo cameras, IMU, Motion capture &
				Stereo cameras, LiDAR, GPS/IMU, Radar &
				Panoramic cameras, Stereo, LiDAR, GPS/IMU \\
				\hline
				\textbf{Unique Features} & 
				Multi-angle views, Controlled scene capture, Transformation metadata & 
				Urban/rural mix, Calibrated sensors, Dense point clouds & 
				Long-term data, Weather variations, Large scale mapping &
				High-precision ground truth, Dynamic motions, Indoor industrial scenes &
				Same routes in spring, summer, fall, winter; Adverse weather &
				360° view, Dense semantic labels, Mesh annotations \\
				\hline
				\textbf{Scene Variety} & 
				High (200+ distinct scenes) & 
				Medium (22 scenes) & 
				Low (same route, different conditions) &
				Low (11 sequences, 3 environments) &
				Medium (28 sequences, same routes in 4 seasons) &
				High (73 sequences in diverse areas) \\
				\hline
				\textbf{Viewpoint Changes} & 
				High (multi-angle captures) & 
				Low (forward-facing camera) & 
				Low (forward-facing camera) &
				Medium (flying patterns) &
				Low (vehicle trajectories) &
				High (360° panoramic views) \\
				\hline
				\textbf{Loop Closure Frequency} & 
				High (designed for loop closure) & 
				Medium (some revisited areas) & 
				High (repeated routes) &
				Low (few loops per sequence) &
				High (same routes in different seasons) &
				High (multiple revisits in urban grid) \\
				\hline
			\end{tabular}%
		}
	\end{table*}
	
	\section{Related Work}
	
	A comparison between LoopDB and the notable datasets used in computer vision research is presented in Table~\ref{table:comparison}. Although LoopDB is smaller (1000 images), it contains a much larger variety of scenes, as opposed to KITTI \cite{geiger2012kitti}, Oxford RobotCar \cite{maddern2017robotcar}, or EuRoC MAV dataset \cite{burri2016euroc}
	
	Loop-closure detection minimizes localization drift and enhances the mapping accuracy \cite{Tsintotas_2022}. Although not designed for benchmarking loop closure, the following datasets were used because of the lack of a better alternative.
	
	The KITTI dataset \cite{geiger2012kitti} is a popular dataset used in the computer vision and robotics communities. This dataset included stereo camera images, LiDAR scans, and GPS measurements gathered by driving a vehicle in urban city environments and rural roads. Although KITTI is not targeted for loop closure detection, it has significantly contributed to SLAM and visual odometry research.
	
	The 4Seasons dataset \cite{wenzel20204seasons} provides multi-weather data across all four seasons in Zurich.What makes 4Seasons special is the systematic coverage of challenging environmental conditions, such as snow, rain, sunshine, and seasonal foliage changes on the same routes, to evaluate visual place recognition and loop closure algorithms under different appearance conditions.
	
	KITTI-360 \cite{liao2022kitti} extends the original KITTI dataset by providing dense 360° semantic and geometric annotations of urban scenes. Collected in Karlsruhe, Germany, includes data from two fisheye cameras, two high-resolution stereo cameras, a Velodyne HDL-64E LiDAR, and GPS/IMU sensors covering approximately 320,000 images and 100,000 laser scans across 73.7 km of urban driving.
	
	The RobotCar dataset contains more than 1000 km of driving data collected in Oxford, UK, for up to a year, which included varying degrees of weather and lighting conditions \cite{maddern2017robotcar}. The data are from a diverse and rich collection of sensors, including stereo cameras, LiDAR, and GPS/INS, which makes it important to test long-term autonomy and vision/place recognition algorithms.
	
	The EuRoC MAV dataset \cite{burri2016euroc} focuses on micro aerial vehicles (MAVs), which are recorded with synchronized stereo images, IMU measurements, and ground truth. While initially designed for the evaluation of visual-inertial (VI) odometry, it has also been broadly utilized in loop closure detection research, most notably for indoor environments.

	\section{LoopDB Dataset}
	
	\begin{figure*}
		\centering
		\includegraphics[width=1\linewidth]{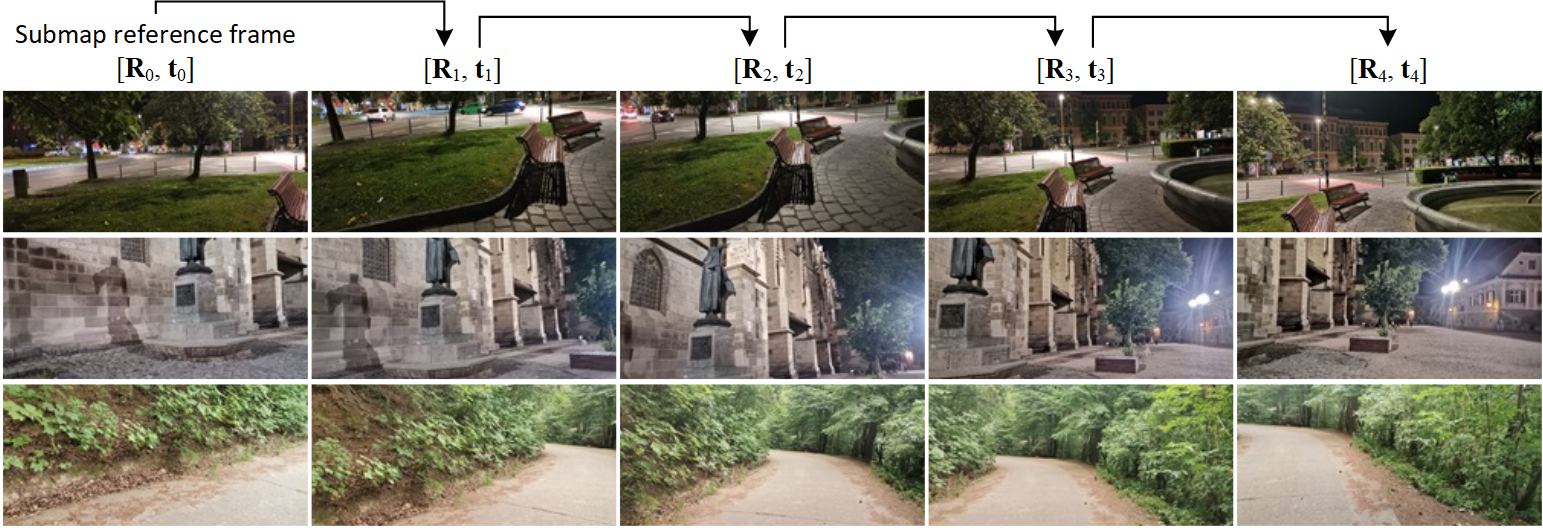}
		\caption{\textbf{Snapshots from the LoopDB dataset (ours).} Each row corresponds to a submap sequence. The rotation and translation $[\vec{R}, \vec{t}]$ between two consecutive images is given as ground truth.}
		\label{fig:dataset}
	\end{figure*}
	
	The LoopDB dataset includes multiple scenes of different environments, such as parking spots, crosswalks, gyms, parks, and institute hallways, among others, captured by a camera sensor from different positions and angles. Along with the images, we publish the rotation and translation $[\vec{R}, \vec{t}]$ between each consecutive image in a sequence. In this work, such a sequence is defined as a submap. A couple of submaps from LoopDB are illustrated in Fig.~\ref{fig:dataset}. 
	
	\subsection{Conventions and Notation}
	
	The dataset  follows the standard conventions commonly used in visual SLAM and loop-closure detection. Images are saved as high-quality jpeg images to show details without excessive storage overhead. We used homogeneous coordinates to represent points in the images, with the notation otherwise standard. This is very important when dealing with transformations such as translation, rotation, and scaling because it allows us to express these operations through matrix multiplication.For example, in 2D space a point $X$ is represented as  $X = \begin{bmatrix} X & Y & 1 \end{bmatrix}^T$, where $X$ and $Y$ are the coordinates of the point in the image plane, and the final component is set to 1 to facilitate affine transformations. Matrices that operate on these points are written as the homography matrix $H$, which encodes the transformations between views.

	\subsection{Submap Homography Transformation}
	
	The key feature of the dataset is the transformed data for each image. As this dataset is designed for loop closure detection and SLAM systems, there is a dire need for accurate transformation data between different views of any scene. Each pair of images has a rotation matrix and translation vector assigned to avoid the need for users to calculate the relative pose between the images themselves. the transformation between two images  captured by homography, denoted by $H$, which is a $3 \times 3$ matrix that encapsulates both the rotation and translation parts. The transformation from the first image coordinates $X_1$ to the second image coordinates $X_2$ is given by: 
	
	\begin{equation}
		X_2 = H \cdot X_1
	\end{equation}
	
	\noindent where $X_1$ and $X_2$ represent points in homogeneous coordinates:
	
	\begin{equation}
		X_1 = \begin{bmatrix} X_1 \\ Y_1 \\ 1 \end{bmatrix}, \quad X_2 = \begin{bmatrix} X_2 \\ Y_2 \\ 1 \end{bmatrix}
	\end{equation}
	
	The homography matrix $H$ contains two parts: the \textit{rotation matrix} $R$ and \textit{translation vector} $t$. In the case of 3D transformations, matrix takes the form:
	
	\begin{equation}
		H = \begin{bmatrix} R & t \\ 0 & 1 \end{bmatrix}
	\end{equation}
	
	\noindent where $R$ is the $3 \times 3$ rotation matrix, $t$ is the $3 \times 1$ translation vector, which describe the movement of the camera between the two images. The translation between images is represented by vector $t = [t_x, t_y, t_z]$.
	
	Instead of directly storing the rotation matrices, we use \textit{quaternions}, since they are compact and avoid gimbal lock issues. A quaternion $q$ is represented as  $q = [q_x, q_y, q_z, q_w]$, where $q_x, q_y, q_z$ are the imaginary components of the quaternion and $q_w$ is the scalar part. Quaternions are used to represent the 3D orientation of the camera and they enable smooth transitions between different orientations.
	
	The full transformation between the two images is expressed as:
	
	\begin{equation}
		\begin{bmatrix} X_2 \\ Y_2 \\ Z_2 \end{bmatrix} = R \cdot \begin{bmatrix} X_1 \\ Y_1 \\ Z_1 \end{bmatrix} + t
	\end{equation}

	\subsection{Data Format}

\begin{figure}[H]
		\centering
		\includegraphics[width=0.5\linewidth]{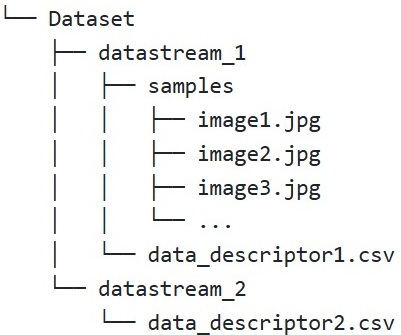}
		\caption{\textbf{LoobDB folder structure.} The data is organized as datastreams of images and their corresponding metadata.}
		\label{dataset_directory}
	\end{figure}

	LoopDB contains images and CSV metadata files organized into two data streams. Each scene (submap) comprised five images with different viewing angles, starting from a root image with zero rotation and translation. The dataset structure is shown in Fig.~\ref{dataset_directory} includes the following.
	
	\begin{enumerate}
		\item \textit{Datastream 1}: Scene images with metadata listing image names and directory paths.
		\item \textit{Datastream 2}: Detailed annotations providing transformation data between images, including root image identification, inter-image transformations, and sequential relationships.
	\end{enumerate}
	
	This organization enables accurate reconstruction of spatial relationships between images, supporting pose estimation, visual odometry, and SLAM applications.%

	\subsection{Metadata Information}
	
	LoopDB is organized using two large metadata files in CSV format for the detailed annotation of each image, as mentioned above. The CSV files provide rotational and transformational information for each image-pair structure. All the rows are mapped to an image corresponding to the image timestamp with a specific rotation (quaternion), translation, or other metadata.
	
	\begin{enumerate}
		\item The metadata in directory \texttt{datastream\_1} contains:
		\begin{itemize}
			\item \textit{timestamp\_start}: timestamp for starting the image capture;
			\item \textit{timestamp\_stop}: timestamp when the image capture is finalized;
			\item \textit{sampling\_time}: time spent for capturing the image;
			\item \textit{left\_file\_path}: directory path of the image in the dataset from the left sensor.
			\item \textit{right\_file\_path}: directory path of the image in the dataset from the right sensor. The current version of LoopDB uses a monocamera, hence only the left sensor being used.
		\end{itemize}
		
		\item The metadata in directory \texttt{datastream\_2} contains:
		\begin{itemize}
			\item \textit{timestamp\_start}: Refers to the image name.
			\item \textit{timestamp\_stop}: Refers to the root image that the image belongs too.
			\item \textit{sampling\_time}: Indicates how many times the same image is captured.
			\item \textit{ timestamp\_root}: Refers to root image name of the scene images collection.
			\item \textit{q\_1 , q\_2 , q\_3 , q\_w}: Refers to the rotation.
			\item \textit{tx , ty , tz}: Refers to the translation.
		\end{itemize}
	\end{enumerate}
	
	\subsection{Software Tools}
	
	The dataset, along with all necessary scripts, is publicly available on GitHub (\url{https://github.com/RovisLab/LoopDB}) ,and the he dataset images are available on Zenodo at (\url{www.doi.org/10.5281/zenodo.15201910}).
	
	Researchers can download the image files and CSV metadata directly or integrate them into their SLAM/loop closure detection pipelines using the provided Python scripts. The repository includes detailed documentation with step-by-step setup instructions, experiment guidelines, and customization options.
	
	To ensure high-quality data processing, a range of software libraries and programming languages were used. Most notably:
	
	\begin{itemize}
		\item \textit{OpenCV}: for image resizing, contrast enhancement, histogram equalization, and feature extraction using SIFT and ORB.
		\item \textit{Python}: as the primary language for dataset management, API development, and testing.
	\end{itemize}
	
	These tools were chosen for their flexibility and widespread use in both computer vision and robotics, ensuring ease of adoption by the research community.

	\section{Scope and Applications}
	
	LoopDB addresses specific challenges in loop closure detection that are critical for robust SLAM systems, supporting both traditional feature-based methods and emerging deep learning approaches. Researchers can use LoopDB in several ways.
	
	\textbf{Benchmarking Traditional Methods:} The dataset provides a standardized test for evaluating feature-based algorithms (SIFT \cite{lowe2004distinctive}, ORB \cite{rublee2011orb}, BRIEF \cite{calonder2010brief}, DAISY \cite{tola2010daisy}) under challenging viewpoint variations.
	
	\textbf{Training Data for Learning-Based Approaches:} With 1,000+ high-quality images across 200+ distinct scenes, LoopDB offers sufficient diversity for training deep learning models, with precise transformation metadata facilitating the creation of positive and negative sample pairs.
	
	\textbf{Transformer-based Methods:} The dataset is well suited for evaluating transformer architectures that leverage attention mechanisms to establish robust correspondences across viewpoint changes. Methods such as MixVPR \cite{ali2023mixvpr} and TransVPR \cite{wang2022transvpr} can be effectively benchmarked using LoopDB multi-angle captures.
	
	\textbf{Advanced Applications:} LoopDB supports geometric verification studies using RANSAC variants or learning-based approaches such as SuperGlue \cite{sarlin2020superglue} and enables the evaluation of modern SLAM systems such as DROID-SLAM \cite{teed2021droid} and DeepFactors \cite{czarnowski2020deepfactors}. The controlled environment of the dataset allows for the isolation and evaluation of loop closure components with precise transformation metadata. By providing a specialized dataset focused on controlled viewpoint variations, LoopDB fills a critical gap in the research ecosystem, enabling the targeted development and evaluation of both traditional and learning-based approaches to this fundamental SLAM challenge.

	\section{Sensors}
	
	Data were captured using a 48MP Huawei Nova 7i camera selected for high-quality imaging across diverse lighting and environmental conditions. Images were taken with varying ISO settings and exposure times to introduce brightness, contrast, and noise variations, which are essential for testing the algorithms under challenging conditions.
	
	Without GPS positioning, we relied on careful camera placement to capture each scene from multiple perspectives. Transformation matrices between consecutive images were computed and stored in metadata files. As shown in Figure~\ref{fig:intro}, each scene is photographed from multiple viewpoints while maintaining a consistent central reference, ensuring coherent image sequences for studying visual continuity across perspectives.
	
	The camera was calibrated for focal length, principal point, and lens distortion to ensure geometrically accurate images for reliable feature matching. Ground-truth alignment was established through precise manual control over camera positioning, with each scene photographed from varying perspectives while maintaining a consistent focal point across all five images in a sequence.

	\subsection{Calibration and Synchronization}
	
	Calibration is essential to correct distortions and ensure accurate camera properties. In this dataset, the camera was calibrated for focal length, principal point, and lens distortion, ensuring geometrically accurate images where features can be reliably matched across different views. Since a single camera was used for data collection, images were manually synchronized to ensure consistent alignment. This approach guarantees that consecutive frames maintain a coherent temporal relationship with minimal warping between them.
	
	\subsection{Ground-Truth Alignment:}
	
	Ground truth alignment was established through precise manual control over the camera position during image capture. Each scene was deliberately photographed from varying perspectives, while maintaining a consistent focal point across all five images in a set. Although this ground truth is based on visual alignment rather than external pose measurements, it is well-suited for SLAM applications, particularly for monocular visual systems that do not require external positioning data.

	\section{Experimental Evaluation}

	To validate LoopDB, we conducted feature-correspondence experiments using four descriptors (SIFT \cite{lowe2004distinctive}, ORB \cite{rublee2011orb}, DAISY \cite{tola2010daisy}, and BRIEF \cite{calonder2010brief}) to assess the projection error. 
	The key findings include:
	\textbf{Distance Effect:} Error increases with distance from reference images.
	\textbf{Spatial Patterns:} Errors show consistent spatial correlation patterns.
	\textbf{Texture Importance:} Well-textured regions produce better feature correspondences.
	\textbf{Viewpoint Challenges:} Substantial errors ($>$100px in some cases) demonstrate dataset complexity.
	
	We evaluated the mean reprojection error across transitions in our dataset (Fig.~\ref{fig:dataset}). As shown in Table \ref{tab:combined_projection_error}, BRIEF performed best (7.47px mean error), followed by DAISY (14.76px), with ORB and SIFT showing higher errors (30.66px and 35.95px). The 1→2 transition was most challenging, with errors ranging from 14.64px (BRIEF) to 163.51px (SIFT). Table \ref{tab:combined_projection_error} confirms BRIEF's superior performance across all transitions.
	
	\begin{table}[H]
		\caption{Reprojection error across transitions and methods.}
		\label{tab:combined_projection_error}
		\centering
		\footnotesize 
		\setlength{\tabcolsep}{5pt}

		\begin{tabular}{|c|c|c|c|c|c|c|}
			\hline
			\multirow{2}{*}{\textbf{Method}} & \multicolumn{4}{c|}{\textbf{Transition Pairs (Error/Match)}} & \multicolumn{2}{c|}{\textbf{Overall}} \\
			\cline{2-7}
			& \textbf{0→1} & \textbf{1→2} & \textbf{2→3} & \textbf{3→4} & \textbf{Avg} & \textbf{Max} \\
			\hline
			SIFT & \begin{tabular}[c]{@{}c@{}}4.16\\731\end{tabular} & \begin{tabular}[c]{@{}c@{}}163.5\\454\end{tabular} & \begin{tabular}[c]{@{}c@{}}8.35\\591\end{tabular} & \begin{tabular}[c]{@{}c@{}}3.71\\1052\end{tabular} & 35.9 & 163.5 \\
			\hline
			ORB & \begin{tabular}[c]{@{}c@{}}10.2\\679\end{tabular} & \begin{tabular}[c]{@{}c@{}}121.8\\767\end{tabular} & \begin{tabular}[c]{@{}c@{}}12.4\\777\end{tabular} & \begin{tabular}[c]{@{}c@{}}9.0\\958\end{tabular} & 30.7 & 121.8 \\
			\hline
			Daisy & \begin{tabular}[c]{@{}c@{}}1.42\\490\end{tabular} & \begin{tabular}[c]{@{}c@{}}65.3\\281\end{tabular} & \begin{tabular}[c]{@{}c@{}}5.30\\436\end{tabular} & \begin{tabular}[c]{@{}c@{}}1.82\\847\end{tabular} & 14.8 & 65.3 \\
			\hline
			BRIEF & \begin{tabular}[c]{@{}c@{}}8.39\\770\end{tabular} & \begin{tabular}[c]{@{}c@{}}14.6\\690\end{tabular} & \begin{tabular}[c]{@{}c@{}}8.71\\723\end{tabular} & \begin{tabular}[c]{@{}c@{}}5.60\\1025\end{tabular} & 7.5 & 14.6 \\
			\hline
		\end{tabular}
	\end{table}
	
	We also compared LoopDB with five popular datasets by using the same descriptors (Table \ref{tab:cross_dataset_comparison}). LoopDB showed three key advantages:
	\textbf{High Feature Density} - extracting significantly more features (SIFT: 810, BRIEF: 679, DAISY: 688);.
	\textbf{Detector Versatility}, which performs well with both SIFT (0.78px) and binary descriptors; and \textbf{Geometric Consistency}, maintaining competitive error rates despite extracting many more features than other datasets. These results establish LoopDB as an effective benchmark for loop closure detection methods, with its 5-frame chain structure enabling the evaluation of algorithm performance across sequential transformations, simulating error accumulation in practical SLAM applications.
	
	\begin{table}[H]
		\caption{Cross-Dataset Comparison of Mean Reprojection Errors (pixels)}
		\label{tab:cross_dataset_comparison}
		\centering
		\begin{tabular}{|l|c|c|c|c|}
			\hline
			\textbf{Dataset} & \textbf{SIFT} & \textbf{ORB} & \textbf{DAISY} & \textbf{BRIEF} \\
			\hline
			LoopDB (Ours) & 0.78 & 44.67 & 5.64 & 4.46 \\
			\hline
			KITTI \cite{geiger2012kitti} & 1.16 & 33.87 & 1.24 & 1.82 \\
			\hline
			Oxford \cite{maddern2017robotcar} & 1.32 & 33.70 & 5.47 & 10.64 \\
			\hline
			EuRoC \cite{burri2016euroc} & 0.56 & 0.53 & 0.79 & 0.79 \\
			\hline
			4Seasons \cite{wenzel20204seasons} & 10.28 & 35.42 & 0.81 & 0.93 \\
			\hline
			KITTI-360 \cite{liao2022kitti} & 8.62 & 100.58 & 1.11 & 1.97 \\
			\hline
		\end{tabular}
	\end{table}

	\section{Conclusions}
	
	This study introduces a challenging and diverse dataset specifically designed for loop closure detection and SLAM. The dataset includes over 1,000 high-quality images captured in a variety of environments, such as parks, gyms, and parking areas. It also provides multiple views of each scene, along with corresponding transformation data, making it a valuable resource for training and testing autonomous navigation algorithms.
	
	Future enhancements may include integrating additional sensors such as GPS and IMU to improve pose estimation and ground truth alignment. Moreover, introducing dynamic elements, such as moving objects and varying weather conditions, would further increase the dataset’s realism. These improvements would allow SLAM and loop closure detection algorithms to be tested in highly dynamic environments—an essential step toward advancing robust and adaptable autonomous systems.

	\subsection{Future Work}
	
	We plan to expand LoopDB to make the dataset more useful for both the traditional and learning-based methods. Our strategy includes: \textbf{Environmental Diversity:} We will add more challenging locations, such as busy city areas, industrial complexes, and underground structures. \textbf{Condition Variations:} We'll capture sequences under challenging weather conditions like rain, snow, and fog, with varying illumination (dawn, dusk, night), and seasonal changes. \textbf{Dynamic Elements:} Include sequences of moving objects and different crowd sizes. \textbf{Trajectory Complexity:} Longer sequences with complex trajectories are added, including multiple loop closures of different sizes within a single sequence. 
	
	The expanded dataset will support deep learning applications with over 7,000 images to enable effective training of neural networks for loop closure detection.

	\bibliographystyle{IEEEtran}
	\bibliography{references}
	
\end{document}